\documentclass[11pt,a4paper]{article}
\usepackage[hyperref]{acl2020}
\usepackage{times}
\usepackage{latexsym}
\usepackage{amsfonts}
\usepackage{booktabs}
\usepackage{multirow}

\newcommand{\B}[1]{\mathbf{#1}}

\newcommand{\Cal}[1]{\mathcal{#1}}
\newcommand{\Sref}[1]{\S\ref{#1}}
\mathchardef\mhyphen="2D

\usepackage{microtype}

\aclfinalcopy 
\def\aclpaperid{1916} 

\title{Topological Sort for Sentence Ordering}

\author{Shrimai Prabhumoye, Ruslan Salakhutdinov, Alan W Black \\
  School of Computer Science\\
  Carnegie Mellon University \\
  Pittsburgh, PA, USA \\
  \texttt{\{sprabhum, rsalakhu, awb\}@cs.cmu.edu} \\}

\date{}

\begin{document}
\maketitle
\begin{abstract}
Sentence ordering is the task of arranging the sentences of a given text in the correct order.
Recent work using deep neural networks for this task has framed it as a sequence prediction problem.
In this paper, we propose a new framing of this task as a constraint solving problem and introduce a new technique to solve it.
Additionally, we propose a human evaluation for this task.
The results on both automatic and human metrics across four different datasets show that this new technique is better at capturing coherence in documents.
\end{abstract}

\section{Introduction}

Sentence ordering is the task of arranging sentences into an order which maximizes the coherence of the text \cite{barzilay2008modeling}.
This is important in applications where we have to determine the sequence of pre-selected set of information to be presented.
This task has been well-studied in the community due to its significance in down stream applications such as ordering of: concepts in concept-to-text generation \cite{konstas2012concept}, information from each document in multi-document summarization \cite{barzilay2002inferring, nallapati2017summarunner}, events in storytelling \cite{fan2019strategies, hu2019makes}, cooking steps in recipe generation \cite{chandu-etal-2019-storyboarding}, and positioning of new information in existing summaries for update summarization \cite{prabhumoye-etal-2019-towards}.
Student essays are evaluated based on how coherent and well structured they are.
Hence, automated essay scoring \cite{burstein2010using, miltsakaki2004evaluation} can use this task to improve the efficiency of their systems.

Early work on coherence modeling and sentence ordering task uses probabilistic transition model based on vectors of linguistic features \cite{lapata2003probabilistic}, content model which represents topics as states in an HMM \cite{barzilay2004catching}, and entity based approach \cite{barzilay2008modeling}.
Recent work uses neural approaches to model coherence and to solve sentence ordering task.
\citet{li2014model} introduced a neural model based on distributional sentence representations using recurrent or recursive neural networks and avoided the need of feature engineering for this task.
In \cite{li2017neural}, they extend it to domain independent neural models for coherence and they introduce new latent variable Markovian generative models to capture sentence dependencies.
These models used windows of sentences as context to predict sentence pair orderings.
\citet{gong2016end} proposed end-to-end neural architecture for sentence ordering task which uses pointer networks to utilize the contextual information in the entire piece of text.

Recently hierarchical architectures have been proposed for this task. 
In \cite{logeswaran2018sentence}, the model uses two levels of LSTMs to first get the encoding of the sentence and then get the encoding of the entire paragraph.
\citet{cui2018deep} use a transformer network for the paragraph encoder to allow for reliable paragraph encoding.
Prior work \cite{logeswaran2018sentence, cui2018deep, kumar2019deep} has treated this task as a sequence prediction task where the order of the sentences is predicted as a sequence.
The decoder is initialized by the document representation and it outputs the index of sentences in sequential order.
Only in \cite{chen2016neural}, this task is framed as a ranking problem.
In this work, a pairwise score is calculated between two sentences and then the final score for an order is obtained by summing over all the scores between pairs of sentences.
The order which has the maximum score is given as output.
Instead of considering all possible permutations of a given order, it uses beam-search strategy to find a sub-optimal order.

Most of the recent work \cite{gong2016end, logeswaran2018sentence, cui2018deep} tries to leverage the contextual information but has the limitation of predicting the entire sequence of the order.
This has the drawback that the prediction at the current time step is dependent on the prediction of the previous time step.
Another limitation of the prior work is the availability of good sentence representations that can help in determining the relative order between two sentences.

For this work we frame the task as a constraint learning problem.
We train a model which learns to predict the correct constraint given a pair of sentences.
The constraint learnt by our model is the relative ordering between the two sentences.
Given a set of constraints between the sentences of a document, we find the right order of the sentences by using sorting techniques.
Since we don't attach a score to an order, we don't have to consider all the permutations of an order.

Our main contribution is a new framing for the sentence ordering task as a constraint solving problem.
We also propose a new and simple approach for this task in this new framework. 
We show that a simple sorting technique can outperform the previous approaches by a large margin given that it has good sentence representations.
The bottleneck for most of the hierarchical models is memory required by the representations of all the sentences and the representation of the paragraph.
The new framing also obviates these memory issues.
The code can be found at \url{https://github.com/shrimai/Topological-Sort-for-Sentence-Ordering}.
Additionally, we introduce a human evaluation for this task and show that our model outperforms the state-of-the-art on all the metrics.

\section{Methodology}
For our task we have a set of $N$ documents $\Cal{D} = \{d_1. \ldots, d_{N}\}$.
Let the number of sentences in each document $d_i$ be denoted by $v_i$, where  $\forall i$, $v_i >= 1$. 
Our task can be formulated as - If we have a set $\{s_{o_1}, \ldots, s_{o_{v_{i}}}\}$ of $v_i$ sentences in a random order where the random order is $\B{o} = [o_1, \ldots, o_{v_{i}}]$, then the task is to find the right order of the sentences $\B{o}^{*} = [o^*_1, \ldots, o^*_{v_{i}}]$.
Prior work \cite{logeswaran2018sentence, cui2018deep} learns to predict the sequence of the correct order $\B{o}^{*}$.
In this formulation of the task, we have $\Cal{C}_i$ set of constraints for document $d_i$.
These constraints $\Cal{C}_i$ represent the relative ordering between every pair of sentences in $d_i$.
Hence, we have $|\Cal{C}_i|  = {v_{i} \choose 2}$.
For example, if a document has four sentences in the correct order $s_1 < s_2 < s_3 < s_4$, then we have six set of constraints $\{s_1 < s_2, s_1 < s_3, s_1 < s_4, s_2 < s_3, s_2 < s_4, s_3 < s_4\}$.
Constraints $\Cal{C}_i$ are learnt using a classifier neural network described in (\Sref{sec:constraint}).
We finally find the right order $\B{o}^*$ using topological sort on the relative ordering between all the $\Cal{C}_i$ pairs of sentences. 

\subsection{Topological Sort}
Topological sort \cite{Tarjan1976} is a standard algorithm for linear ordering of the vertices of a directed graph.
The sort produces an ordering $\B{\hat{o}}$ of the vertices such that for every directed edge $u \rightarrow v$ from vertex $u$ to vertex $v$, $u$ comes before $v$ in the ordering $\B{\hat{o}}$.
We use the depth-first search based algorithm which loops through each node of the graph, in an arbitrary order.
The algorithm visits each node $n$ and prepends it to the output ordering $\B{\hat{o}}$ only after recursively calling the topological sort on all descendants of $n$ in the graph.
The algorithm terminates when it hits a node that has been visited or has no outgoing edges (i.e. a leaf node).
Hence, we are guaranteed that all nodes which depend on $n$ are already in the output ordering $\B{\hat{o}}$ when the algorithm adds node $n$ to $\B{\hat{o}}$. 

We use topological sort to find the correct ordering $\B{o}^*$ of the sentences in a document.
The sentences can represent the nodes of a directed graph and the directed edges are represented by the ordering between the two sentences.
The direction of the edges are the constraints predicted by the classifier.
For example, if the classifier predicts the constraint that sentence $s_1$ precedes $s_2$, then the edge $s_1 \rightarrow s_2$ would be from node of $s_1$ to $s_2$.

This algorithm has time complexity of $O(v_i+|\Cal{C}_i|)$ for a document $d_i$.
In our current formulation, all the constraints are predicted before applying the sort.
Hence, we have to consider all the $|\Cal{C}_i| = {v_i \choose 2}$ edges in the graph.
The time complexity of our current formulation is $O(v_i^2)$.
But the same technique could be adopted using a Merge Sort \cite{DBLP:books/lib/Knuth98a} algorithm in which case the time complexity would be $O(v_i \log v_i)$.
In this case, the sort algorithm is applied first and the constraint is predicted only for the two sentences for which the relative ordering is required during the sort time.

\subsection{Constraint Learning}
\label{sec:constraint}
We build a classifier to predict a constraint between two sentences $s_1$ and $s_2$ (say).
The constraint learnt by the classifier is the relative ordering between the two sentences.
Specifically, the classifier is trained to predict whether $s_2$ follows $s_1$ or not i.e the the classifier predicts the constraint $s_1 < s_2$.

\paragraph{BERT based Representation. (B-TSort)}
We use the Bidirectional Encoder Representations from Transformers (BERT) pre-trained uncased language model \cite{devlin2019bert} and fine-tune it on each dataset using a fully connected perceptron layer.
Specifically, we leverage the Next Sentence Prediction objective of BERT and get a single representation for both sentences $s_1$ and $s_2$.
The input to the BERT model is the sequence of tokens of sentence $s_1$, followed by the separator token `[SEP]', followed by the sequence of tokens for sentence $s_2$.
We use the pooled representation for all the time steps\footnote{This code was based on \cite{Wolf2019HuggingFacesTS}.}.
\vspace{-0.25em}
\paragraph{LSTM based Representation. (L-TSort)}
In this model we get two separate representations $\B{h_1}$ and $\B{h_2}$ for $s_1$ and $s_2$ from a bi-directional LSTM encoder, respectively.
We pass the concatenation of $\B{h_1}$ and $\B{h_2}$ as input to two layers of perceptron for constraint prediction.
This model is trained to gain insight on the contribution of pre-trained sentence representations for the constraint prediction formulation of the task.


\section{Experimental Results}
\label{sec:length}


This section describes the datasets, the evaluation metric and the results of our experiments. The hyper-paramater settings are reported in Apendix.


\subsection{Datasets}

\paragraph{NSF. NIPS, AAN abstracts.} These three datasets contain abstracts from NIPS papers,
ACL papers, and the NSF Research Award Abstracts dataset respectively and are introduced in \cite{logeswaran2018sentence}. 
The paper also provides details about the statistics and processing steps for curating these three datasets.


\paragraph{SIND caption.} We also consider the SIND (Sequential Image Narrative Dataset) caption dataset \cite{huang2016visual} used in the sentence ordering task by \cite{gong2016end}.
All the stories in this dataset contain five sentences each and we only consider textual stories for this task.


\subsection{Baselines}

\paragraph{Attention Order Network (AON).} 
This is the current state-of-the-art model \cite{cui2018deep} which formulates the sentence ordering task as a order prediction task.
It uses a LSTM based encoder to learn the representation of a sentence.
It then uses a transformer network based paragraph encoder to learn a representation of the entire document.
It then decodes the sequence of the order by using a LSTM based decoder.

\vspace{-0.25em}

\paragraph{BERT Attention Order Network (B-AON).}
To have a fair comparison between our model and the AON model, we replace the LSTM based sentence representation with the pre-trained uncased BERT model.
This model plays a pivotal role of giving us an insight into how much improvement in performance we get only due to BERT.

\vspace{-0.25em}

\subsection{Evaluation Metric}

\paragraph{Perfect Match (PMR):} calculates the percentage of samples for which the entire sequence was correctly predicted \cite{chen2016neural}. $\mathsf{PMR} = \frac{1}{N} \sum_{i=1}^{N} 1\{\B{\hat{o}}^{i} = \B{o}^{*i}\}$, where $N$ is the number of samples in the dataset.
It is the strictest metric.

\vspace{-0.25em}

\paragraph{Sentence Accuracy (Acc):} measures the percentage of sentences for which their absolute position was correctly predicted \cite{logeswaran2018sentence}.
$\mathsf{Acc} = \frac{1}{N} \sum_{i=1}^{N} \frac{1}{v_i} \sum_{j=1}^{v_{i}} 1\{\B{\hat{o}}_{j}^{i} = \B{o}_{j}^{*i}\}$
, where $v_i$ is the number of sentences in the $i^{th}$ document.
It is a also a stringent metric.

\vspace{-0.25em}

\paragraph{Kendall Tau (Tau):} quantifies the distance between the predicted order and the correct order in terms of the number of inversions \cite{lapata2006automatic}.
$\tau = 1 - 2 I / {v_i \choose 2}$, where $I$ is the number of pairs in the predicted order with incorrect relative order and $\tau \in [-1, 1]$.

\vspace{-0.25em}

\paragraph{Rouge-S:} calculates the percentage of skip-bigrams for which the relative order is predicted correctly \cite{chen2016neural}.
Skip-bigrams are the total number of pairs $v_i \choose 2$ in a document.
Note that it does not penalize any arbitrary gaps between two sentences as long as their relative order is correct.
$\mathtt{Rouge\mhyphen S} = \frac{1}{{v_i \choose 2}} \mathtt{Skip}(\B{\hat{o}}) \cap \mathtt{Skip}(\B{o}^*)$
, where the $\mathtt{Skip}(.)$ function returns the set of skip-bigrams of the given order.

\vspace{-0.25em}

\paragraph{Longest Common Subsequence (LCS):} calculates the ratio of longest common sub-sequence \cite{gong2016end} between the predicted order and the given order (consecutiveness is not necessary, and higher is better).

\vspace{-0.25em}

\paragraph{Human Evaluation}
We introduce a human evaluation experiment to assess the orders predicted by the models.
We set up a manual pairwise comparison following \cite{bennett2005large} and present the human judges with two orders of the same piece of text.
The judges are asked ``Pick the option which is in the right order according to you.''
They can also pick a third option `No Preference' which corresponds to both the options being equally good or bad.
In total we had 100 stories from the SIND dataset\footnote{We choose SIND because all the stories contain 5 sentences and hence it is easy to read for the judges. The orders of the stories are easier to judge as compared to the orders of scientific abstracts like NSF, NIPS and AAN as they require the judges to have an informed background.} annotated by 10 judges.
We setup three pairwise studies to compare the B-TSort vs AON order, B-TSort vs Gold order and AON vs Gold order (Gold order is the actual order of the text).
Each judge annotated a total of 30 stories, 10 in each of the above mentioned categories. 
The judges were naive annotators.


\subsection{Results}
\begin{table}[t]
\centering
\small{
\begin{tabular}{@{}l r r r r r@{}}
\textbf{Model} & \textbf{PMR} & \textbf{Acc} & \textbf{Tau} & \textbf{Rouge-S} & \textbf{LCS} \\
\toprule
\multicolumn{6}{c}{\textbf{NIPS abstracts}} \\
\midrule
AON & 16.25 & 50.50 & 0.67 & 80.97 & 74.38 \\
B-AON & 19.90 & 55.23 & 0.73 & 83.65 & 76.29 \\
L-TSort & 12.19 & 43.08 & 0.64 & 80.08 &  71.11 \\
B-TSort & \textbf{32.59} & \textbf{61.48} & \textbf{0.81} & \textbf{87.97} & \textbf{83.45} \\
\toprule
\multicolumn{6}{c}{\textbf{SIND captions}} \\
\midrule
AON & 13.04 & 45.35 & 0.48 & 73.76 & 72.15 \\
B-AON & 14.30 & 47.73 & 0.52 & 75.77 & 73.48 \\
L-TSort & 10.15 & 42.83 & 0.47 & 73.59 & 71.19 \\
B-TSort & \textbf{20.32} & \textbf{52.23} & \textbf{0.60} & \textbf{78.44} & \textbf{77.21} \\
\toprule
\end{tabular}
}
\vspace{-1.0em}
\caption{
\label{nips_sind}
Results on NIPS and SIND datasets
}
\vspace{-1.0em}
\end{table}

Table \ref{nips_sind} shows the results of the automated metrics for the NIPS and SIND datasets\footnote{We fine-tune BERT which is memory intensive. Hence, we show the results of B-AON only on these two datasets as they need 2 transformer layers for paragraph encoder \cite{cui2018deep}}.
It shows that AON\footnote{We use the code provided by the authors to train the AON and B-AON model. The numbers reported in Table \ref{nips_sind} and \ref{nsf_acn} are our runs of the model. Hence, they differ from the numbers reported in the paper \cite{cui2018deep}.}  model gains on all metrics when the sentence embeddings are switched to BERT.
The L-TSort model which does not utilize BERT embeddings comes close to AON performance on Rouge-S and Tau metrics. 
This demonstrates that the simple L-TSort method is as accurate as AON in predicting relative positions but not the absolute positions (PMR and Acc metric).
Table \ref{nips_sind} shows that our method B-TSort does not perform better only due to BERT embeddings but also due to the design of the experiment.
Note that BERT has been trained with the Next Sentence Prediction objective and not the sentence ordering objective like ALBERT \cite{Lan2020ALBERT:}.
We believe that framing this task as a constraint solving task will further benefit from pre-trained language model like ALBERT.
Table \ref{nsf_acn} shows results for the NSF and AAN datasets and the B-TSort model performs better than the AON model on all metrics.

\begin{table}[t]
\centering
\small{
\begin{tabular}{@{}l r r r r r@{}}
\textbf{Model} & \textbf{PMR} & \textbf{Acc} & \textbf{Tau} & \textbf{Rouge-S} & \textbf{LCS} \\
\toprule
\multicolumn{6}{c}{\textbf{NSF abstracts}} \\
\midrule
AON & 13.18 & 38.28 & 0.53 & 69.24 & 61.37 \\
B-TSort & 10.44 & 35.21 & 0.66 & 69.61 & 68.50 \\
\toprule
\multicolumn{6}{c}{\textbf{AAN abstracts}} \\
\midrule
AON & 36.62 & 56.22 & 0.70 & 81.52 & 79.06 \\
B-TSort & 50.76 & 69.22 & 0.83 & 87.76 & 85.92 \\
\toprule
\end{tabular}
}
\vspace{-0.6em}
\caption{
\label{nsf_acn}
Results on NSF and AAN datasets
}
\end{table}

\begin{table}
\centering
\small{
\begin{tabular}{@{}r r r@{}}
\toprule
\textbf{B-TSort} & \textbf{No Preference} & \textbf{B-AON} \\
\midrule
\textbf{41.00}\% & 28.00\% & 31.00\% \\
\toprule
\textbf{B-TSort} & \textbf{No Preference} & \textbf{Gold} \\
\midrule
26.00\% &  20.00\% & \textbf{54.00}\% \\
\toprule
\textbf{B-AON} & \textbf{No Preference} & \textbf{Gold} \\
\midrule
24.00\% & 22.00\% & \textbf{54.00}\% \\
\toprule
\end{tabular}
}
\vspace{-0.6em}
\caption{
\label{human_eval}
Human Evaluation Results on B-TSort vs AON (top), B-TSort vs Gold (middle) and AON vs Gold (bottom).
}
\vspace{-1.5em}
\end{table}

Table \ref{human_eval} shows results for the three human evaluation studies on the SIND dataset.
It shows that human judges prefer B-TSort orders 10\% more number of times than the B-AON orders\footnote{Examples of B-TSort and B-AON orders are shown in Table \ref{tab:sind} and \ref{tab:nips} for SIND and NIPS dataset in Appendix.}. 
The reference order may not be the only correct ordering of the story.
The variability in the orders produced by B-TSort and B-AON is not very high and hence in comparison with Gold orders, we don't see much difference in human preferences.

\begin{table*}[t]
    \centering
    \small{
    \begin{tabular}{@{}l @{\hskip 2em} r r r r @{\hskip 2em} r r r r @{}}
        \toprule
        \textbf{Model}  & \textbf{Win=1} & \textbf{Win=2} & \textbf{Win=3} & \textbf{\% Miss} & \textbf{Win=1} & \textbf{Win=2} & \textbf{Win=3} & \textbf{\% Miss} \\
        \midrule
        & \multicolumn{4}{c@{\hskip 2em}}{\textbf{NIPS}} & \multicolumn{4}{c}{\textbf{SIND}}\\
        \cmidrule(lr{1.8em}){2-5} \cmidrule(lr{0.2em}){6-9} B-AON & 81.81 & 92.44 & 96.50 & 3.48 & 78.39 & 92.79 & 98.43 & 0.00 \\
        B-TSort & 87.59 & 95.59 & 98.11 & 0.00 & 82.67 & 95.01 & 99.09 & 0.00 \\
        \midrule
        & \multicolumn{4}{c@{\hskip 2em}}{\textbf{NSF}} & \multicolumn{4}{c}{\textbf{AAN}}\\
        \cmidrule(lr{1.8em}){2-5} \cmidrule(lr{0.2em}){6-9} AON & 50.58 & 63.87 & 72.96 & 5.85 & 82.65 & 92.25 & 96.73 & 0.84 \\
        B-TSort & 61.41 & 75.52 & 83.87 & 0.00 & 90.56 & 96.78 & 98.71 & 0.00 \\
    \bottomrule
    \end{tabular}
    }
    \vspace{-0.6em}
    \caption{
    \label{tab:displacement-analysis}
    Sentence Displacement Analysis for all the datasets. (Win=Window size; \% Miss=\% mismatch) 
    }
    \vspace{-1.0em}
\end{table*}

The low scores of AON could be due to the fact that it has to decode the entire sequence of the order.
The search space for decoding is very high (in the order of $v_{i}!$).
Since our framework, breaks the problem to a pairwise constraint problem, the search space for our model is in the order of $v^2_i$.



\paragraph*{Discussion:} 
We perform additional analysis to determine the displacement of sentences in the predicted orders of the models, scalability of the models for longer documents, and an understanding of quality of the human judgements.

Displacement of sentences in predicted orders is measured by calculating the percentage of sentences whose predicted location is within 1, 2 or 3 positions (in either direction) from their original location. 
A higher percentage indicates less displacement of sentences.
We observed that in spite of lack of a global structure, B-TSort consistently performs better on all datasets for all three window sizes as shown in Table~\ref{tab:displacement-analysis}.
Observe that as window size reduces, the difference between B-TSort and B-AON percentages increases. 
This implies that displacement of sentences is higher in B-AON despite taking the whole document into account. 

We additionally perform a comparison of models on documents containing more than 10 sentences and the results are shown in Table~\ref{tab:struct-long}. 
B-TSort consistently performs better on all the metrics. 
SIND dataset is omitted in these experiments as the maximum number of sentences in the story is five for all the stories in the dataset.
For each dataset, the Tau difference for longer documents is much higher than the Tau difference on the overall dataset (Table~\ref{nips_sind} and \ref{nsf_acn}). 
This implies that B-TSort performs much better for longer documents.

Note that the AON model generates the order and hence need not generate positions for all the sentences in the input. 
We calculate the percentage of mismatches between the length of the input document and the generated order. 
For AON model on the NSF dataset which has longest documents, the overall mismatch is 5.85\% (Table~\ref{tab:displacement-analysis}), while the mismatch for documents with more than 10 sentences is 11.60\%.
The AON model also produces an overall mismatch of 0.84
\% on AAN documents while producing a mismatch of 5.17\% on longer AAN documents.
Similarly, the B-AON model has an overall mismatch of 3.48\% for NIPS dataset, and 33.33\% mismatch for longer documents.
This problem does not arise in our design of the task as it does not have to stochastically generate orders. 

To better understand the choices of human judges, we observe the average length of stories calculated in number of tokens.
For the B-TSort vs B-AON study, we discover that the average length of the stories for B-TSort, B-AON and `No Preference' chosen options is 86, 65 and 47 respectively. 
This means that B-TSort is better according to human judges for longer stories. 
Similarly for B-TSort vs Gold experiment, the human judges were confused with longer stories, reiterating that B-TSort performs well with long stories. 

\begin{table}[t]
    \centering
    \small{
    \begin{tabular}{@{}l @{\hskip 2em} r r r r r @{}}
        \toprule
        \textbf{Model} & \textbf{PMR} & \textbf{Acc} & \textbf{Tau} & \textbf{Rouge-S} & \textbf{LCS} \\
        \midrule
            & \multicolumn{5}{c@{\hskip 2em}}{\textbf{NIPS abstracts}}\\
        \cmidrule(l{-0.2em}r{0.2em}){2-6} B-AON & 0.0 & 29.18 & 0.51 & 74.64 & 63.81 \\
        B-TSort & 0.0 & 39.43 & 0.74 & 83.26 & 71.68 \\
        \midrule
        & \multicolumn{5}{c@{\hskip 2em}}{\textbf{NSF abstracts}}\\
        \cmidrule(l{-0.2em}r{0.2em}){2-6} AON & 2.12 & 21.42 & 0.41 & 67.45 & 55.47 \\
        B-TSort & 0.67 & 28.57 & 0.64 & 68.46 & 64.86 \\
        \midrule
        & \multicolumn{5}{c@{\hskip 2em}}{\textbf{AAN abstracts}}\\
        \cmidrule(l{-0.2em}r{0.2em}){2-6} AON & 0.0 & 22.70 & 0.40 & 68.90 & 56.19 \\
        B-TSort & 0.0 & 36.86 & 0.69 & 78.52 & 72.01 \\
        \bottomrule
    \end{tabular}
    }
    \vspace{-0.6em}
    \caption{
    \label{tab:struct-long}
    Analysis on NIPS, NSF and AAN datasets for documents longer than 10 sentences.
    }
    \vspace{-1.5em}
\end{table}

\section{Conclusion and Future Work}
We have shown a new way to design the task of sentence ordering.
We provide a simple yet efficient method to solve the task which outperforms the state of the art technique on all metrics.
We acknowledge that our current model has the limitation of not including the entire context of the paragraph while making the decision of the relative order of the pairs.
Our future work is to include the paragraph representation in the constraint prediction model.
This will help our methodology to have the benefit of making informed decision while also solving constraints.

\section*{Acknowledgments}
This work was supported in part by ONR Grant N000141812861, NSF IIS1763562, and Apple.
We would also like to acknowledge NVIDIA’s GPU support.

\bibliography{acl2020}
\bibliographystyle{acl_natbib}

\appendix
\twocolumn

\section{Appendix}
\paragraph{Hyper-parameters. } 
For AON model we use the code base provided by the authors in \cite{cui2018deep} and we maintain the hyper-parameters described in the paper. 
For the paragraph encoder of the B-AON models, we follow the same scheme of the AON model but for its sentence encoder we use hyper-parameters of the BERT setting.
We use the pretrained BERT uncased base model with 12 layers for the B-AON and B-TSORT models. 
We fine-tune the BERT model in both cases.
Hence, we replace the Adadelta optimizer with the BertAdam \cite{Wolf2019HuggingFacesTS} optimizer for the B-AON model.
The LSTMs in the L-TSort model uses an RNN size of 512 and it uses the same vocabularies as the AON model.
L-TSort is trained using stochastic gradient descent with dropout of 0.2, learning rate of 1.0 and learning decay rate of 0.5. 
For B-TSort and L-TSort we use accuracy on the validation set to stop training.
For B-TSort and B-AON we use learning rate of 5e-5 with adam epsilon value of 1e-8.
For all the experiments we use a maximum sequence length of 105 tokens.

\begin{table*}[t]
\centering
\setlength{\tabcolsep}{0.25em} 
\small{
\begin{tabular}{p{1.7in}@{\hskip 0.3in} p{1.7in}@{\hskip 0.3in} p{1.7in}} 
\toprule
\multicolumn{1}{c}{\textbf{Gold Order}}{\hskip 0.3in} & \multicolumn{1}{c}{\textbf{B-TSort Order}}{\hskip 0.3in} & \multicolumn{1}{c}{\textbf{B-AON Order}}  \\ 
\midrule
 \multicolumn{3}{c}{SIND Dataset}       \\ 
\midrule
the family sits together for dinner on the first night of the annual reunion. the restaurant we chose had amazing food and everyone loved the presentation. gemma really adored the restaurants decorations and was always gazing at them. aunt harriot had a little trouble deciding what kind of wine she wanted tonight. bob had the whole family cracking up with his jokes.  & the family sits together for dinner on the first night of the annual reunion. the restaurant we chose had amazing food and everyone loved the presentation. aunt harriot had a little trouble deciding what kind of wine she wanted tonight. gemma really adored the restaurants decorations and was always gazing at them. bob had the whole family cracking up with his jokes. & the family sits together for dinner on the first night of the annual reunion. aunt harriot had a little trouble deciding what kind of wine she wanted tonight. bob had the whole family cracking up with his jokes. gemma really adored the restaurants decorations and was always gazing at them. the restaurant we chose had amazing food and everyone loved the presentation.  \\ 
\addlinespace
he wanted to take a ride on his new bike. we went on a nice ride out to the lake. we really enjoyed the beautiful view from the dock. it was very peaceful watching the boats. we had such a busy day he needed a nap.  & we went on a nice ride out to the lake. he wanted to take a ride on his new bike. we really enjoyed the beautiful view from the dock. it was very peaceful watching the boats. we had such a busy day he needed a nap. & we went on a nice ride out to the lake. he wanted to take a ride on his new bike. it was very peaceful watching the boats. we really enjoyed the beautiful view from the dock. we had such a busy day he needed a nap. \\
\addlinespace
when we finally brought our son home from the hospital so many people were at home with us to see him. everyone wanted a chance to hold him! we were all so happy to have a new addition to the family. my parents were so proud to be grand parents! i am so happy and i love my son very much!  & when we finally brought our son home from the hospital so many people were at home with us to see him. we were all so happy to have a new addition to the family. everyone wanted a chance to hold him! my parents were so proud to be grand parents! i am so happy and i love my son very much! & my parents were so proud to be grand parents! when we finally brought our son home from the hospital so many people were at home with us to see him. we were all so happy to have a new addition to the family. everyone wanted a chance to hold him! i am so happy and i love my son very much! \\ 
\midrule
\end{tabular}
}
\vspace{-0.8em}
\caption{\centering Examples of predicted sentence orders for B-TSort and B-AON model for SIND dataset.}
\vspace{-0.5em}
\label{tab:sind}
\end{table*}

\begin{table*}[t]
\centering
\setlength{\tabcolsep}{0.25em} 
\small{
\begin{tabular}{p{1.9in}@{\hskip 0.2in} p{1.9in}@{\hskip 0.2in} p{1.9in}} 
\toprule
\multicolumn{1}{c}{\textbf{Gold Order}}{\hskip 0.2in} & \multicolumn{1}{c}{\textbf{B-TSort Order}}{\hskip 0.2in} & \multicolumn{1}{c}{\textbf{B-AON Order}}  \\ 
\midrule
 \multicolumn{3}{c}{NIPS Dataset}       \\ 
\midrule
we study how well one can recover sparse principal components of a data matrix using a sketch formed from a few of its elements. we show that for a wide class of optimization problems, if the sketch is close (in the spectral norm) to the original data matrix, then one can recover a near optimal solution to the optimization problem by using the sketch. in particular, we use this approach to obtain sparse principal components and show that for m data points in n dimensions, o(-2k max{m, n}) elements gives an - additive approximation to the sparse pca problem (k is the stable rank of the data matrix). we demonstrate our algorithms extensively on image, text, biological and financial data. the results show that not only are we able to recover the sparse pcas from the incomplete data, but by using our sparse sketch, the running time drops by a factor of five or more. & 
we study how well one can recover sparse principal components of a data matrix using a sketch formed from a few of its elements. we show that for a wide class of optimization problems, if the sketch is close (in the spectral norm) to the original data matrix, then one can recover a near optimal solution to the optimization problem by using the sketch. in particular, we use this approach to obtain sparse principal components and show that for m data points in n dimensions, o(-2k max{m, n}) elements gives an - additive approximation to the sparse pca problem (k is the stable rank of the data matrix). the results show that not only are we able to recover the sparse pcas from the incomplete data, but by using our sparse sketch, the running time drops by a factor of five or more. we demonstrate our algorithms extensively on image, text, biological and financial data. & 
we study how well one can recover sparse principal components of a data matrix using a sketch formed from a few of its elements. in particular, we use this approach to obtain sparse principal components and show that for m data points in n dimensions, o(-2k max{m, n}) elements gives an - additive approximation to the sparse pca problem (k is the stable rank of the data matrix). we show that for a wide class of optimization problems, if the sketch is close (in the spectral norm) to the original data matrix, then one can recover a near optimal solution to the optimization problem by using the sketch. the results show that not only are we able to recover the sparse pcas from the incomplete data, but by using our sparse sketch, the running time drops by a factor of five or more. we demonstrate our algorithms extensively on image, text, biological and financial data. \\
\addlinespace
we develop a latent variable model and an efficient spectral algorithm motivated by the recent emergence of very large data sets of chromatin marks from multiple human cell types . a natural model for chromatin data in one cell type is a hidden markov model ( hmm ) ; we model the relationship between multiple cell types by connecting their hidden states by a fixed tree of known structure . the main challenge with learning parameters of such models is that iterative methods such as em are very slow , while naive spectral methods result in time and space complexity exponential in the number of cell types . we exploit properties of the tree structure of the hidden states to provide spectral algorithms that are more computationally efficient for current biological datasets . we provide sample complexity bounds for our algorithm and evaluate it experimentally on biological data from nine human cell types . finally , we show that beyond our specific model , some of our algorithmic ideas can be applied to other graphical models . &
a natural model for chromatin data in one cell type is a hidden markov model ( hmm ) ; we model the relationship between multiple cell types by connecting their hidden states by a fixed tree of known structure . the main challenge with learning parameters of such models is that iterative methods such as em are very slow , while naive spectral methods result in time and space complexity exponential in the number of cell types . we develop a latent variable model and an efficient spectral algorithm motivated by the recent emergence of very large data sets of chromatin marks from multiple human cell types . we exploit properties of the tree structure of the hidden states to provide spectral algorithms that are more computationally efficient for current biological datasets . we provide sample complexity bounds for our algorithm and evaluate it experimentally on biological data from nine human cell types . finally , we show that beyond our specific model , some of our algorithmic ideas can be applied to other graphical models . &
the main challenge with learning parameters of such models is that iterative methods such as em are very slow , while naive spectral methods result in time and space complexity exponential in the number of cell types . a natural model for chromatin data in one cell type is a hidden markov model ( hmm ) ; we model the relationship between multiple cell types by connecting their hidden states by a fixed tree of known structure .', 'we develop a latent variable model and an efficient spectral algorithm motivated by the recent emergence of very large data sets of chromatin marks from multiple human cell types . we exploit properties of the tree structure of the hidden states to provide spectral algorithms that are more computationally efficient for current biological datasets . we provide sample complexity bounds for our algorithm and evaluate it experimentally on biological data from nine human cell types . finally , we show that beyond our specific model , some of our algorithmic ideas can be applied to other graphical models . \\
\midrule
\end{tabular}
}
\vspace{-0.8em}
\caption{\centering Examples of predicted sentence orders for B-TSort and B-AON model for NIPS dataset.}
\vspace{-0.5em}
\label{tab:nips}
\end{table*}

\end{document}